\documentclass[lettersize,journal]{IEEEtran}
\usepackage{amsmath,amsfonts}
\usepackage{algorithmic}
\usepackage{algorithm}
\usepackage{array}
\usepackage[caption=false,font=normalsize,labelfont=sf,textfont=sf]{subfig}
\usepackage{textcomp}
\usepackage{stfloats}
\usepackage{url}
\usepackage{verbatim}
\usepackage{graphicx}
\usepackage{cite}
\usepackage{xcolor}
\usepackage{times}
\usepackage{helvet}
\usepackage{courier}
\usepackage{colortbl}
\newtheorem{definition}{Definition}

\newtheorem{theorem}{Theorem}
\usepackage{multirow}
\usepackage{amssymb}
\usepackage{engord}
\usepackage{booktabs}
\usepackage{hyperref}
\begin{document}

\title{Differential Privacy Personalized Federated Learning Based on Dynamically Sparsified Client Updates}

\author{Chuanyin Wang \textsuperscript{1,2,3}, 
Yifei Zhang\textsuperscript{1},
Neng Gao\textsuperscript{1},
Qiang Luo\textsuperscript{4}\\
\thanks{Corresponding author: Yifei Zhang (zhangyifei@iie.ac.cn)} 
\IEEEmembership{\textsuperscript{1}Institute of Information Engineering, Chinese Academy of Sciences, Beijing, 100085, China\\
\textsuperscript{2}Key Laboratory of Cyberspace Security Defense, Beijing, 100085, China\\
\textsuperscript{3}School of Cyber Security, University of Chinese Academy of Sciences, Beijing, 100049, China\\
\textsuperscript{4}No.208 Research Institute of China Ordnance Industries, Beijing, 102202, China\\
\textsuperscript{} Email: wangcyy520@163.com, zhangyifei@iie.ac.cn, gaoneng@iie.ac.cn, hongtdhlb@163.com}}





\maketitle
\begin{abstract}
Personalized federated learning is extensively utilized in scenarios characterized by data heterogeneity, facilitating more efficient and automated local training on data-owning terminals. This includes the automated selection of high-performance model parameters for upload, thereby enhancing the overall training process. However, it entails significant risks of privacy leakage. Existing studies have attempted to mitigate these risks by utilizing differential privacy. Nevertheless, these studies present two major limitations: (1) The integration of differential privacy into personalized federated learning lacks sufficient personalization, leading to the introduction of excessive noise into the model. (2) It fails to adequately control the spatial scope of model update information, resulting in a suboptimal balance between data privacy and model effectiveness in differential privacy federated learning. In this paper, we propose a differentially private personalized federated learning approach that employs dynamically sparsified client updates through reparameterization and adaptive norm(DP-pFedDSU). Reparameterization training effectively selects personalized client update information, thereby reducing the quantity of updates. This approach minimizes the introduction of noise to the greatest extent possible. Additionally, dynamic adaptive norm refers to controlling the norm space of model updates during the training process, mitigating the negative impact of clipping on the update information. These strategies substantially enhance the effective integration of differential privacy and personalized federated learning. Experimental results on EMNIST, CIFAR-10, and CIFAR-100 demonstrate that our proposed scheme achieves superior performance and is well-suited for more complex personalized federated learning scenarios. 
\end{abstract}
\newenvironment{notetopractioners}{
  \renewcommand{\abstractname}{Note to Practitioners}  
  \begin{abstract}  
}{
  \end{abstract}  
}

\begin{notetopractioners}

The main objective of personalized federated learning is to enable clients with heterogeneous data to effectively utilize their local data to train a global model, without sharing their raw data. However, the method of sharing model parameters in personalized federated learning still poses a risk of privacy leakage, and existing research employing differential privacy for protection does not fully achieve the desired outcome. Specifically, the introduction of differential privacy significantly reduces the effectiveness of the global model. Therefore, the primary aim of this paper is to efficiently balance data privacy and model effectiveness in personalized federated learning. By employing our proposed solution, it is ensured that, under the same privacy budget, the algorithm can achieve higher model accuracy. Furthermore, the method can further enhance the personalization of local models in this paper, allowing each participating client to select model parameter information specific to their local data, which can better strengthen the application of federated learning in scenarios with data heterogeneity. In future research, we will focus on refining the balance between model effectiveness and data privacy based on the proposed approach.

\end{notetopractioners}


\section{Introduction}
\IEEEPARstart{F}{ederated} Learning \cite{mcmahan2017communication} aims to collaboratively train models across multiple clients without sharing local private data. Although federated learning avoids sharing local data, the interactive data during the training is not absolutely secure. An attacker can infer the client's data features or membership based on interaction information (such as gradients) \cite{fredrikson2015model, melis2019exploiting, nasr2019comprehensive}. Moreover, attackers can reconstruct the client's private data from the gradients \cite{hitaj2017deep, lam2021gradient, zhu2019deep}, leading to severe privacy breaches for the client.

Differentially private federated learning  \cite{el2022differential} has become a widely adopted approach for protecting data privacy in federated learning . Differential privacy \cite{dwork2014algorithmic} reduces the sensitivity of shared information in federated learning by clipping updates and adding noise to the communicated data. Existing differentially private federated learning algorithms focus on protecting local data instances \cite{agarwal2018cpsgd, hu2023federated} and securing interactive information \cite{cheng2022differentially, geyer2017differentially, mcmahan2017learning}.

Research indicates that real-world data distributions are highly heterogeneous, which has led to the development of personalized federated learning  \cite{collins2021exploiting, pillutla2022federated, oh2021fedbabu}. The aforementioned differentially private federated learning algorithms, primarily based on traditional federated learning methods, face challenges when applied to personalized federated learning due to data heterogeneity. To address these challenges, researchers have proposed differentially private federated learning algorithms tailored for personalized federated learning \cite{shen2023share, jain2021differentially}. In these algorithms, the model is divided into a feature extraction part and a classification part. Only the feature extraction component is exchanged between clients and the server, and differential privacy is applied solely to this portion. While this approach reduces the amount of noise added to the model, its effectiveness is limited.

Moreover, much of the existing research focuses on enhancing differential privacy mechanisms or improving the robustness of federated learning algorithms to integrate these paradigms effectively. However, these approaches often overlook a crucial aspect: not all communicated information in federated learning is equally important or sensitive. Therefore, it may not be necessary to apply differential privacy uniformly across all model parameters. Some studies propose sparsifying the uploaded parameters to reduce the volume of communicated data \cite{cheng2022differentially, qu2022generalized}. However, these methods typically involve selecting parameters after training, which may inadvertently discard valuable information essential for the model's performance.

\begin{figure*}[!t]
\centering
\includegraphics[scale=0.6]{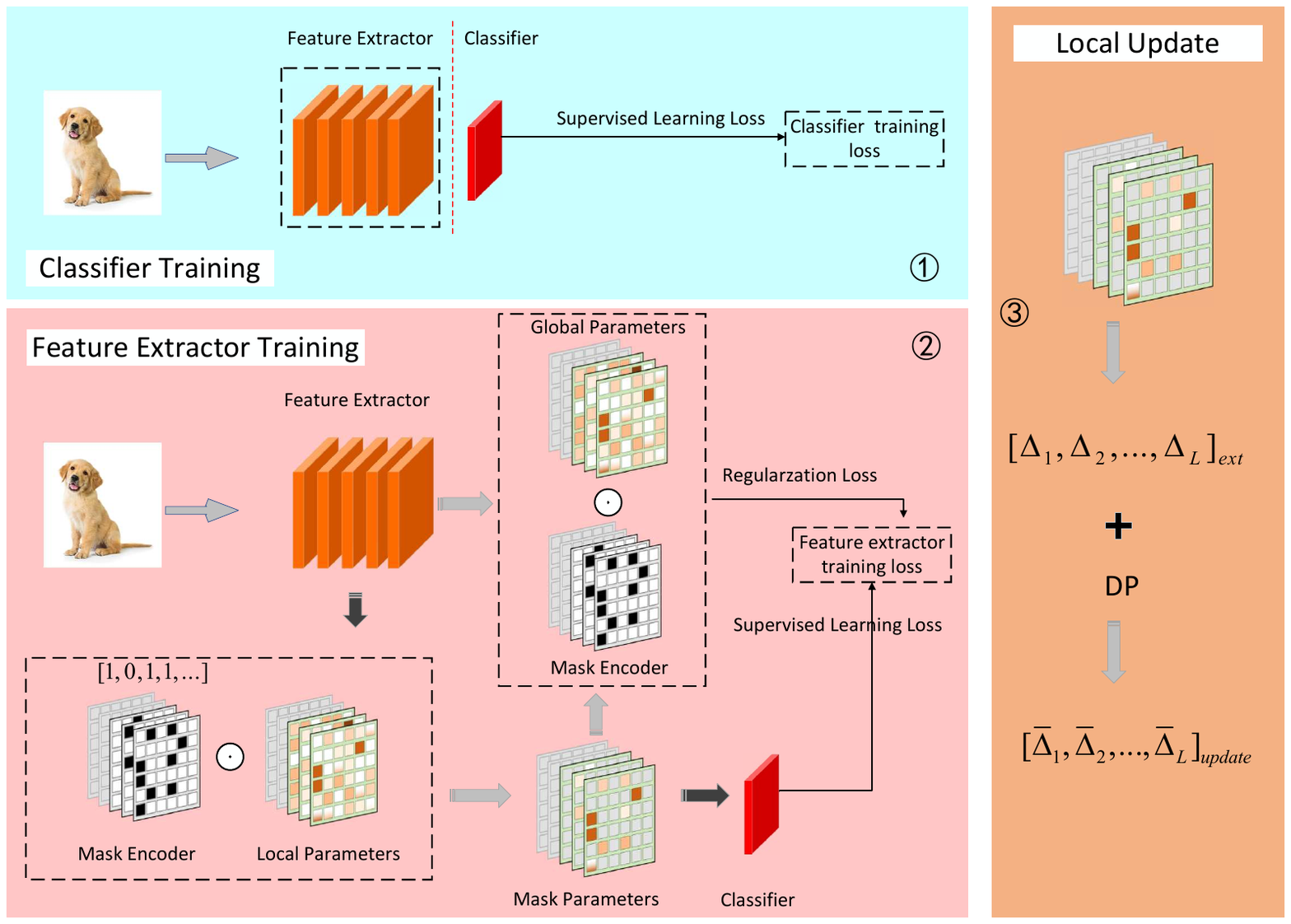}
\caption{\centering The overflow of client training in the proposed method.}
\label{figure:1}
\end{figure*}

To address these challenges, we propose a differentially private personalized federated learning method based on dynamically sparsified client updates, which leverages reparameterization and adaptive norms. Our approach aims to more effectively filter interactive information in personalized federated learning by selecting and protecting only the most important and sensitive data. Firstly, we decompose the model into two components: a feature extractor and a classifier. The parameters of the feature extractor are used to compute the interaction information exchanged between the server and the clients, while the classifier parameters are retained locally on each client. Secondly, we employ reparameterization during model training to encourage useful information to concentrate within a specific subset of parameters determined by a predefined sparsity level. After training, only this subset of parameters is retained and transmitted. This strategy reduces the volume of interactive information, thereby introducing less noise under the same privacy guarantees while maintaining model performance. Additionally, we introduce a regularization term to ensure that the norm of the interaction information after reparameterization training remains within a specified range. This helps mitigate the performance degradation caused by the clipping of interaction information during differential privacy operations.

The main contributions of this paper are:
\begin{itemize}
    \item We propose an efficient differential privacy personalized federated learning approach that implements model training using reparameterization.
    \item We introduce a regularization term to control the norm of interactive information, effectively reducing the impact of clipping on model performance.
    \item We conduct experiments on multiple datasets, demonstrating that our approach outperforms the state-of-the-art under the same privacy guarantees.
\end{itemize}

\begin{table*}
    \renewcommand{\arraystretch}{1.5}
    \centering
    \caption{Main notation description in our method.}
    \setlength{\tabcolsep}{5mm}{
    \begin{tabular}{c|c}
        \toprule[1.5pt]
        Notation & Description\\
        \midrule[1pt]
        $\epsilon$ & Privacy budget\\
        $\mathcal{Sen}$ & Sensitivity\\
        T & The number of global rounds\\
        $\tau$ & The number of local iterations \\
        $\eta_{l}$ & The local learning rates of active agents\\
        $\eta_{g}$ & The global learning rates of server\\
        $\theta$ & Model parameters\\
        $\theta_{ext}$ & The parameters of the feature extractor in client model partitioning for personalized federated learning\\
        $\theta_{cls}$ & The parameter of classifier in client model partitioning for personalized federated learning\\
        $\Delta$ & The update information from each client after completing local training\\
        $B$ & The batch-size of each local iteration\\
        $\Omega$ & The  active agents of each global round\\
        $\sigma$ & Noise scale  of differential privacy\\
        $C$ & The clipping threshold of differential privacy\\
        N & Number of agents participating in training\\
        $\varphi$ & The feature extractor in the client model partitioning for personalized federated learning.\\*
        h & The classifier in the client model partitioning for personalized federated learning\\*
        $\lambda$ & The coefficient of the regularization term\\

        \bottomrule[1.5pt]
    \end{tabular}}
    \\ [2mm]
    \label{table_0}
    
\end{table*}

\section{Related Work}
Differential Privacy Federated Learning (DPFL) has become a prominent algorithm for ensuring the secure and efficient operation of federated learning. DP-FedAvg \cite{mcmahan2017learning}, a pioneering algorithm, combines the Gaussian mechanism with composition privacy theory to secure federated learning interaction data. The AE-DPFL algorithm \cite{zhu2020voting} substitutes the conventional gradient averaging process with a voting mechanism on data labels. This approach ensures differential privacy for the data labels uploaded by clients, thereby providing robust user-level privacy guarantees. DP-SCAFFLOD \cite{noble2022differentially} integrates DP with SCAFFLOD and demonstrates the convergence of the DP-SCAFFLOD algorithm for both convex and non-convex objectives. PRIVATEFL \cite{yang2023privatefl}, considering client data heterogeneity, proposes a personalized transformation of local client data to reduce the heterogeneity of data distribution, thus compensating for the heterogeneity introduced by differential privacy. Additionally, \cite{cheng2022differentially} investigates the factors contributing to model performance degradation in federated learning under differential privacy, particularly focusing on the norm of update information and the associated noise. To address these issues, the study introduces Bounded Local Update Regularization, which aims to constrain the norm of local model update parameters and mitigate induced noise through parameter sparsification techniques. Similarly, \cite{hu2023federated} ensures model performance using parameter sparsification. Existing research shows that Sharpness-Aware Minimization (SAM) leads to a min-max optimization problem, in which gradient descent can be efficiently performed\cite{foret2020sharpness}. The study \cite{shi2023make} employs SAM to mitigate the impact of differential privacy on model performance. All these studies are based on traditional federated learning algorithms and lack feasibility analysis in personalized federated learning. Therefore, \cite{yang2023dynamic} proposes dynamically updating global model parameters for each client during every global training round, allowing clients to retain personalized local model parameters. However, this approach requires uploading all local model parameters after training, which poses a higher privacy risk compared to uploading only part of the model parameters in personalized federated learning.

Personalized Federated Learning (PFL) leverages the heterogeneity of data distribution by conducting additional adaptive training on local data \cite{fallah2020personalized, collins2021exploiting, pillutla2022federated, oh2021fedbabu}. Consequently, algorithms ensuring differential privacy in personalized federated learning have also been proposed. The PPSGD algorithm \cite{bietti2022personalization} introduces an additional personalized model that is retained locally. This approach effectively balances model effectiveness and data privacy. However, its generalization guarantee depends on the convexity of the loss function. Accordingly, the CENTAUR algorithm proposed in \cite{shen2023share} ensures convergence for more general non-convex objectives. This algorithm adds differential privacy guarantees only to the shared global representation, with locally retained classifier information unaffected by differential privacy. \cite{xu2023learning} proposed an algorithm similar to \cite{shen2023share}, which utilizes multiple virtual clients to jointly train a global embedding model. While these algorithms combine personalized federated learning and differential privacy, they do not fundamentally balance data privacy and model efficacy. This paper proposes a reparameterization-based approach to obtain personalized shared model parameters, reducing the number of uploaded model parameters while ensuring model performance and effectively safeguarding data privacy. Additionally, a new loss term is introduced to control the norm of shared parameters, thereby mitigating the model efficacy loss caused by clipping operations.

\section{Preliminary}
\subsection{Federated Learning}
In a federated learning system comprising $N$ clients and a central server, the training data is distributed across these $N$ clients. The central server functions as a coordinating hub, tasked with disseminating the global model to each client and subsequently aggregating the local models received from them. The clients are responsible for training their respective local models using their locally available data. Let $\mathcal{D}_{i}$ be the local data of the $i$-th client. If $f(.;\theta)$ is the training model, the local objective for client i can be defined as follows:

\begin{equation}
    l(\theta;\mathcal{D}_{i}) = \frac{1}{\parallel \mathcal{D}_{i} \parallel} \sum_{j \in \mathcal{D}_{i}} \ell (f(x_{ij};\theta),y_{ij}).
\end{equation}

where $(x_{ij}, y_{ij})$ represents the $j$-th sample of the $i$-th client. Therefore, the global objective is defined as:
\begin{equation}
    arg \mathop{\min}_{\theta}\mathcal{L}(\theta) = \frac{1}{N}\sum^{N}_{i=1}l(\theta;\mathcal{D}_{i}).
\end{equation}

Traditional federated learning demonstrates strong performance when data distribution is relatively uniform. However, training data often exhibits significant heterogeneity in practical scenarios, resulting in substantial discrepancies between the outcomes of local optimizations and the ideal global training objective. As a result, differential privacy federated learning also tends to perform suboptimally in heterogeneous federated learning environments.

\subsection{Differential Privacy}
Differential Privacy \cite{dwork2014algorithmic} aims to prevent the disclosure of individual data during data analysis. Due to its strong theoretical foundation, it is widely used in various privacy-preserving scenarios. It is defined as follows:
\begin{definition}
    \textbf{Differential Privacy.} A randomized algorithm $\mathcal{M}$ satisfies ($\epsilon, \delta$)-differential privacy if for any  adjacent datasets $D, D^{'}$ and any subsets of outputs $\mathcal{O} \subseteq \mathcal{M}$, we have that
    \begin{equation}
        Pr(\mathcal{M}(D) \in \mathcal{O}) \leq e^{\epsilon}Pr(\mathcal{M}(D^{'}) \in \mathcal{O})+\delta
    \end{equation}
\end{definition}

Traditional differential privacy has a significant negative impact on the performance of federated learning. Therefore, we utilize a more relaxed version of differential privacy: Renyi Differential Privacy \cite{mironov2017renyi}.
\begin{definition}
    \textbf{Renyi Differential Privacy (RDP).} Given a real number $\alpha \in (1,\infty)$ and privacy parameter $\rho \geq 0$, a randomized algorithm $\mathcal{M}$ satisfies ($\alpha, \rho$)-RDP if for any  adjacent datasets $D, D^{'}$ that differ in a single record,  the Renyi divergence between $\mathcal{M}(\mathcal{D})$ and $\mathcal{M}(\mathcal{D}^{'})$ satisfies:
    \begin{equation}
        D_{\alpha}\left[ \mathcal{M}(\mathcal{D}) \parallel \mathcal{M}(\mathcal{D}^{'}) \right] := \frac{1}{\alpha-1} log \mathbb{E} \left[ \left( \frac{\mathcal{M}(\mathcal{D})}{\mathcal{M}(\mathcal{D}^{'})}\right)^{\alpha} \right] \leq \rho
    \end{equation}
\end{definition}

One of the operations of differential privacy is to add noise to the output of a function $\mathcal{F}$. The scale of the noise depends on the sensitivity of the function $\mathcal{F}$. The sensitivity is defined as follows:

\begin{definition}
    \textbf{$l_{2}$ Sensitivity.} Let $\mathcal{F}$ be a function, the $L_{2}$-sensitivity of  $\mathcal{F}$ defined as $\mathcal{Sen} = max_{D\simeq D^{'}}\parallel\mathcal{F}(D)-\mathcal{F}(D^{'})\parallel_{2}$, where the maximization is taken over all pairs of adjacent datasets.
\end{definition}

\section{Methodology}
Our goal is to achieve better model performance with equal privacy guarantees. We divide the global model parameters $\theta$ into two parts, 
 $[\theta_{ext},\theta_{cls}]$. The parameterized model can be expressed as $f = h \circ \varphi$, where $\varphi$ represents the feature extractor, which extracts data information from different distributions into a common feature space, and $h$ represents the conversion of the data in the feature space into specific labels. Based on the above analysis, the federated learning optimization problem in this paper is transformed into the following form:
\begin{equation}
\begin{aligned}
    &\mathop{\min}_{\theta_{ext}} \frac{1}{N} \sum_{i=1}^{N} \mathop{\min}_{\theta_{i,cls}}  l([\theta_{ext},\theta_{i,cls}];\mathcal{D}_{i})  \\&=  \mathop{\min}_{\theta_{ext}} \frac{1}{N} \sum_{i=1}^{N} \mathop{\min}_{\theta_{i,cls}} \left(\frac{1}{\parallel {D}_{i} \parallel} \sum_{j \in {D}_{i}} \ell(h(\varphi(x_{ij};\theta_{ext});\theta_{i,cls}),y_{ij})\right).
\end{aligned}
\end{equation}

While personalized federated learning can reduce the noise introduced by differential privacy, the resulting improvement in model performance remains limited. Existing methods employ sparsification techniques to further reduce the amount of interaction information, thereby decreasing the noise added to the model \cite{hu2023federated, shi2023make}. Although this approach significantly enhances the performance of traditional differentially private federated learning algorithms, its effectiveness in personalized federated learning is limited. This limitation may stem from the fact that, after receiving the feature extractor information from the server, personalized federated learning fine-tunes the model using local data, which diminishes the benefits of sparsification. Furthermore, the sparsification methods in these studies often select update information using a fixed proportion, potentially removing parameters that are crucial to the model's performance.

However, according to the analysis in \cite{hitaj2017deep}, parameter sparsification can enhance the robustness of the model. Therefore, this paper conducts an in-depth analysis of sparsification techniques and proposes the following two research objectives:
\begin{itemize}
    \item Ensure that the sparsified parameters uploaded by the client contain all important information to the greatest extent possible.
    \item Minimize the decline in model performance caused by the use of differential privacy clipping operations on the client side.
\end{itemize}

Our personalized federated learning algorithm differs from traditional methods in the client-side training, as shown in \hyperref[figure:1]{Fig.~\ref*{figure:1}}. Differentially private personalized federated learning is divided into two main parts. First, each client keeps the parameters of the feature extractor unchanged and trains the local classifier using local data. Second, each client freezes the parameters of the local classifier and trains the parameters of the feature extractor. The distinction of this approach lies in the training process of the feature extractor, where we apply reparameterization to perform sparsified training on the later layers of the model (i.e., the network layers in the gray section are not involved in the sparsification training), combined with dynamic adaptive norm constraints to limit the range of updates. Ultimately, these two approaches reduce the impact of differential privacy on the local updates uploaded by each client.

\subsection{Reparameterization Training}
In personalized federated learning, only the parameters of the feature extractor are transmitted between the server and clients. Therefore, the reparameterized portion focuses on the feature extractor. This study uses parameter sparsification techniques to achieve sparse parameter updates at each client. The objective of reparameterization is to ensure that the final sparse updates retain the most critical information of the model to the greatest extent possible. The reparameterization here is implemented as a forward pass sparsification, ensuring that the weights used in each forward pass are sparse.

Based on the above analysis, we need to maintain a sparse set of model parameters in each round of training. Specifically, the parameters of the feature extractor, $\theta_{ext}$, require a corresponding mask matrix $M$. If we set a sparsification rate $S$, representing the proportion of non-zero parameters in the feature extractor, then $1 \ - S$ represents the proportion of zero parameters. We define the mask matrix $M$ as follows:

\begin{equation}
    M = 
    \left\{
    \begin{array}{ll}
        1 & \text{if } \quad  m \in \text{Topk}(\theta_{ext}^{t}, S) \\
        0 & \text{otherwise}
    \end{array}
    \right.
\end{equation}

Where t denotes the t-th global training round in federated learning. Topk refers to selecting parameters based on their magnitude within each layer, ensuring that the parameters retained each round contribute the most to the feature extractor, as proven in \cite{jayakumar2020top}. Therefore, the actual model parameters for each round are:
\begin{equation}
    \overline \theta_{i,ext}^{t,s} = \theta_{i,ext}^{t,s} \odot M^{s}
\end{equation}

The notation $\theta_{i,ext}^{t,s}$ represents the parameters of the feature extractor for the i-th client in the $s$-th round of local training, while $\overline \theta_{i,ext}^{t,s}$ represents the parameters of the feature extractor's forward propagation for the i-th client in the $s$-th round of local training. The above equation indicates that only the most important parameters for the model are retained each round. Consequently, after several rounds of local updates, the retained model parameters represent the optimal parts for model performance.

Additionally, considering the intrinsic characteristics of neural networks, the shallow layers primarily capture global model features, whereas the deeper layers are more reflective of personalized features. This distinction is pivotal for the realization of personalized federated learning. Accordingly, this paper further refines the reparameterization algorithm by selectively reparameterizing only the deeper layers of the network. This strategy not only preserves the global model's essential characteristics but also optimally retains parameters pertinent to local personalized data. Assuming there are $L$ layers in the network, we reparameterize the last k layers of the feature extractor. The formula is given as:
\begin{equation}
    \overline{\theta}_{i,\text{ext}}^{t,s} = \left\{
    \begin{array}{ll}
    \theta_{i,\text{ext}_j}^{t,s}, & \text{if } 0 < j < k \\
    \theta_{i,\text{ext}_j}^{t,s} \odot M_{j}^{s}, & \text{otherwise}
    \end{array} \right.
\end{equation}

Where $\theta_{i,\text{ext}_j}^{t,s}$ denotes the parameters of the j-th layer of the feature extractor for the i-th client in local training round $s$, while $M_{j}^{s}$ represents the mask value of the j-th layer in the mask matrix corresponding to the feature extractor. Through the above expression, we retain all parameter information for layers below layer k in the feature extractor, while parameters for layer k and above are selected according to the representation in the mask matrix. This approach effectively mitigates the information loss caused by re-parameterization in the shallow layers of the feature extractor. Additionally, this operation significantly reduces the parameter count in higher layers, achieving our goal of sparsification.

\subsection{Dynamic Adaptive Norm}
One effective strategy for implementing differential privacy is clipping, which reduces data sensitivity by uniformly regularizing the magnitude of updates. However, this approach often results in a significant decline in model performance. To address the adverse effects of clipping, prior studies \cite{cheng2022differentially, yang2023dynamic} have proposed controlling the norm space of the updates within a specified clipping threshold range to minimize its impact. Nonetheless, these methods constrain the norm space of the entire parameter set, resulting in a lack of flexibility. In response, this paper introduces a dynamic adaptive norm space restriction method.

To achieve this goal, we introduce a regularization term to limit the norm space of the parameters. Our loss function is as follows:
\begin{equation}
\begin{aligned}
    \mathcal{L} &= \frac{1}{\parallel {D}_{i} \parallel} \sum_{j \in {D}_{i}} \ell(h(\varphi(x_{ij};\theta_{i,ext}^{t,s});\theta_{i,cls}),y_{ij}) \\&+ \frac{\lambda}{2} \Big\| \| \theta_{i,ext}^{t,s}-\theta_{i,ext}^{t,0} \|_{2} -C\Big\|_{2}
\end{aligned}
\end{equation}
Here, ${D}_{i}$ represents the user's private data, $\parallel {D}_{i} \parallel$ represents the number of client's private data, $\theta_{i,ext}^{t,0}$ represents the feature extractor parameters received by the client from the server in the t-th global training round, and $\theta_{i,ext}^{t,s}$ denotes the feature extractor parameters after $s$ rounds of local training. The final updated information of this study is the difference between the trained model parameters and the global model parameters received from the server, $\theta_{i,ext}^{t,\tau}-\theta_{i,ext}^{t,0}$. Therefore, it is necessary to keep its norm size within the clipping threshold $C$.

Considering that the model updates ultimately uploaded are not all model parameters, we optimized the above loss function as follows:
\begin{equation}
\begin{aligned}
    \mathcal{L} &= \frac{1}{\parallel {D}_{i} \parallel} \sum_{j \in {D}_{i}} \ell(h(\varphi(x_{ij};\theta_{i,ext}^{t,s});\theta_{i,cls}),y_{ij}) \\&+ \frac{\lambda}{2} \Big\| \parallel (\theta_{i,ext}^{t,s}-\theta_{i,ext}^{t,0})\cdot M_{i}^{s} \parallel_{2} -C\Big\|_{2}
\end{aligned}
\end{equation}
We modified the calculation method of the newly added regularization term. Originally, it calculates the L2 norm of the difference between the current local round parameters and the initially received global model parameters. In addition to calculating the difference, this difference is multiplied by the mask matrix $M^{s}$ for the corresponding training round. This method not only maximally preserves the performance of the model itself but also dynamically adjusts the norm space of the update information.

\begin{figure}[!t]
    \label{alg1}
    \begin{algorithm}[H]
        \caption{Differential Privacy Personalized Federated Learning Based on Dynamically Sparsified Client Updates(DP-pFedDSU)}
        \begin{algorithmic}
            \REQUIRE Initial model $[\theta_{ext}^{0}, \theta_{cls}^{0}]$, number of global rounds $T$, number of local iterations $\tau$, local learning rates $\eta_{cls}, \eta_{ext}$, global learning rates $\eta_{g}$, number of clients $N$, client sampling probability $q \in (0,1]$, clipping hyperparameter $C$, noise scale $\sigma$   
            \ENSURE Trained feature extractor $\theta^{T}_{ext}$  
        \end{algorithmic}
        \textbf{Server}
        \begin{algorithmic}[1]
            \FOR{$t=1$ to $T-1$}
            \STATE $\Omega_{t} \leftarrow$ Sample clients with probability $q$ over $N$
            \FOR{each client $i \in \Omega_{t}$ in parallel}
            \STATE $\Delta^{t}_{i,ext} = LocalUpdate(\theta^{t}_{ext},i)$;
            \ENDFOR
            \STATE $\theta^{t}_{ext} = \theta^{t-1}_{ext} + \eta_{g}\frac{1}{\| \Omega_{t} \|} \sum_{i \in \Omega_{t}} \Delta^{t}_{i,ext} $;
            \ENDFOR
            \RETURN $\theta^{T}_{ext}$

        \end{algorithmic}
        \textbf{LocalUpdate}
        \begin{algorithmic}[1]
            \STATE $\theta^{t,0}_{i,ext} \leftarrow$ Download $\theta^{t-1}_{i,ext}$
            \STATE Phase 1: Local classifier fine tuning. 
            
            \STATE Phase 2: Local feature extractor update.
            \FOR{$s=0$ to $\tau$ }
            \STATE Sample batch $B\subset \mathcal{D}_{i}$;
            \STATE Compute mask matrix by (6), 
            \STATE Update the parameters of feature extractor by (8), 
            \STATE Compute the loss by (10),
            \ENDFOR
            \STATE Compute local updates $\Delta^{t}_{i,ext} = \theta_{i, ext}^{t,\tau} - \theta_{i,ext}^{t,0} \cdot M^{\tau}$
            \STATE Compute clipped updates of local feature extractor $\widetilde{\Delta}_{i,ext}^{t} = \Delta^{t}_{i,ext}\cdot \min \left(1, \frac{C}{\parallel \Delta^{t}_{i,ext} \parallel_{2}}\right)$ 
            \STATE Add Gaussian noise $\overline{\Delta}^{t}_{i,ext} = \widetilde{\Delta}^{t}_{i,ext} + \mathcal{N}(0,\sigma^{2}C^{2}\cdot \mathbf{I}_{d}/qN)\cdot M^{\tau}$
            
            \RETURN $\overline{\Delta}^{t}_{i,ext}$
        \end{algorithmic}
    \end{algorithm}
\end{figure}

The detailed training steps are shown in Algorithm 1. After the clients complete the training of the feature extractor, they begin to compute local update information. 

By applying the training methods outlined in this section, we achieve a sparsified model that retains only the parameters most critical to the model's performance. Consequently, the update information uploaded by each client should include only these retained model parameters. In this study, we define the interactive information as the difference between the local model and the global model. The final interactive information can be expressed as follows:
\begin{equation}
    \Delta^{t}_{ext} = \theta_{ext}^{t,\tau} - \theta_{ext}^{t,0} \cdot M^{\tau}
\end{equation}
Where $M^{\tau}$ represents the mask matrix after $\tau$ rounds of local updates. Subsequently, differential privacy is employed to protect the update information. First, clipping is applied to reduce the sensitivity of the update information, as shown below:
\begin{equation}
    \widetilde{\Delta}_{ext}^{t} = \Delta\cdot \min \left(1, \frac{C}{\parallel \Delta^{t}_{ext} \parallel_{2}}\right)
\end{equation}
Here, $C$ denotes the clipping threshold. To further enhance data privacy, noise is added to the clipped update information. The final update information after noise addition is expressed as follows:
\begin{equation}
    \overline{\Delta}^{t}_{ext} = \widetilde{\Delta}^{t}_{ext} + \mathcal{N}(0,\sigma^{2}C^{2}\cdot \mathbf{I}_{d}/qN)\cdot M^{\tau}
\end{equation}

\section{Privacy Alalysis}
Compared to DP-FedAvg, this paper employs the Rényi Differential Privacy (RDP) mechanism, which offers tighter privacy bounds. This mechanism is better suited to the distributed training paradigm of federated learning, enabling a more effective balance between model performance and data privacy. During the training process, each client applies differential privacy protection to their update information before uploading it. We utilize the moments accountant method to calculate the cumulative privacy budget throughout training, ensuring that the update information achieves the specified level of privacy protection \cite{abadi2016deep}. Given an algorithm $M$, the moments accountant is defined as follows:
\begin{equation}
\alpha_{\mathcal{M}}(\lambda) \triangleq \max_{\mathcal{D},\mathcal{D}',aux}\log\mathbb{E}[\exp(\lambda c(o;\mathcal{M},aux,\mathcal{D},\mathcal{D}'))]
\end{equation}
where $c(o;\mathcal{M},aux,\mathcal{D},\mathcal{D}^{'})$ represents the privacy loss. The mechanism $\mathcal{M}$ is $(\epsilon, \delta)-DP$ for $\epsilon > 0$ and $ \delta = \min_{\lambda} exp(\alpha_{\mathcal{M}}(\lambda)-\lambda\epsilon)$. Since federated learning involves training with multiple clients, it is necessary to compute a joint privacy budget. Suppose an adaptive mechanism $\mathcal{M}$ is composed of several adaptive mechanisms $\mathcal{M}_{1},...,\mathcal{M}_{k}$. According to the composition rules of the moments accountant, the privacy guarantee of $\mathcal{M}$ can be expressed as:
\begin{equation}
\alpha_{\mathcal{M}}(\lambda) \leq \sum_{i=1}^{k}\alpha_{\mathcal{M}_{i}}(\lambda)
\end{equation}

Combining the above and the privacy theory analysis in \cite{abadi2016deep}, the following privacy guarantee can be derived:
\begin{theorem} Let $q$ denote the sampling rate of participating clients in each round of training in federated learning. Given two constants $c_{1}$ and $c_{2}$, and the number of steps $T$, if $\epsilon < c_{1}q^{2}T$, Algorithm 1 in this paper satisfies $(\epsilon, \delta)$-differential privacy (DP) for any $\delta > 0$, provided that we choose \begin{equation} \sigma \geq c_{2}\frac{q\sqrt{T\log(1/\delta)}}{\epsilon} \end{equation} \end{theorem}

\begin{table*}[ht]
    \renewcommand{\arraystretch}{1.7}
    \centering
    \caption{Model performance comparison of different training methods on multiple datasets.}
    \setlength{\tabcolsep}{5mm}{
    \begin{tabular}{l|c|c|c|c|c|c}
        \toprule[1.5pt]
        \rowcolor{lightgray}
        Method & Number Classes & DP-FedAvg-fb & PPSGD & PMTL-FT & CENTAUR & DP-pFedDSU(Ours)\\
        \midrule[1pt]
        \multirow{2}{*}{EMNIST} 
        &s=5 &69.97 &66.80 &69.97 &81.11 &\textbf{81.80} \\
        &s=10 &54.93 &52.89 &54.63 &68.97 &\textbf{71.87} \\
        \midrule[1pt]
        \multirow{3}{*}{CIFAR-10} & s=2 &83.70 &84.35 &83.34 &85.11 &\textbf{86.29} \\
        &s=5 &49.00 &53.00 &48.86 &53.92 &\textbf{55.76} \\
        &s=10 &36.98 &43.02 &36.29 &41.06 &\textbf{43.92} \\
        \midrule[1pt]

        \multirow{3}{*}{CIFAR-100} 
        &s=3 &47.26 &62.63 &48.26 &60.26 &\textbf{63.47} \\
        &s=5 &29.91 &44.01 &29.67 &43.56 &\textbf{46.00} \\
        &s=10 &17.70 &28.63 &17.58 &28.30 &\textbf{28.80} \\

        \bottomrule[1.5pt]
    \end{tabular}}
    \\ [2mm]
    \label{table_1}
    
\end{table*}

\begin{table*}[ht]
    \renewcommand{\arraystretch}{1.5}
    \centering
    \caption{Comparison of Model Performance Under Different Privacy Budgets.}
    \setlength{\tabcolsep}{15mm}{
    \begin{tabular}{c|c|c|c}
        \toprule[1.5pt]
        \multirow{2}*{Method} & \multicolumn{3}{c}{test accuricy}  \\
        \cmidrule{2-4}
            &$\epsilon$ = 0.5 & $\epsilon$ = 1  & $\epsilon$ = 2 \\
        \midrule[1pt]

        DP-FedAVg-fb &48.84 &49.00 &53.13 \\
        PPSGD &51.62 &53.00 &55.50\\
        PMTL-FT &48.76 &48.86 &52.94 \\
        CENTAUR &53.40 &53.92 &54.85\\
        DP-pFedDSU(Ours) &\textbf{55.71} &\textbf{55.76} &\textbf{55.94}\\
    
        \bottomrule[1.5pt]
    \end{tabular}}
    \\ [2mm]
    \label{table_2}
    
\end{table*}

\begin{table*}[ht]
    \renewcommand{\arraystretch}{1.5}
    \centering
    \caption{Model performance under different data partition scenarios. The $\checkmark$ symbol indicates whether the corresponding component is included.}
    \setlength{\tabcolsep}{11mm}{
    \begin{tabular}{l c|c|c|c}
        \toprule[1.5pt]
        \rowcolor{lightgray}
        \multicolumn{2}{c|}{\textbf{Components}} & \multicolumn{3}{c}{\textbf{Datasets}} \\
        \cmidrule{3-5}
        \rowcolor{lightgray}
        RT&DAN & \textbf{CIFAR-10(10)} & \textbf{CIFAR-10(5)} & \textbf{EMNIST(10)} \\
        \cmidrule{1-5}
        & & 41.06 & 53.92 & 68.97 \\
        \midrule[1pt]
        \checkmark & & 43.38 & 55.05 & 71.10 \\
        \midrule[1pt]
        \checkmark & \checkmark & 43.92 & 55.76 & 71.87 \\
        \bottomrule[1.5pt]
    \end{tabular}}
    \\ [2mm]
    \label{table_3}
\end{table*}

\section{Experiments}
In this section, we test the performance of the proposed experimental scheme on multiple datasets. Additionally, we conduct various comparative experiments to verify the effectiveness of the proposed scheme. Our experiments are conducted on a server equipped with dual Intel Xeon Gold 5218 CPUs(64 cores), 128 GB of memory, and 22.8 TB of storage. The server is also equipped with one NVIDIA A100 GPU, running CUDA 12.2 and PyTorch 1.13.

\subsection{Experimental Setting} \label{setting}

To demonstrate the effectiveness of the proposed scheme, we select state-of-the-art algorithms in personalized federated learning using differential privacy for this experiment.

\begin{itemize}
    \item \textbf{DP-FedAvg-fb}: This algorithm builds on the traditional DP-FedAvg, with the client fine-tuning its classifier head using local data.
    \item \textbf{PPSGD}: This algorithm enhances the personalization performance of the local model using an additive model.
    \item \textbf{PMTL-FT}: This algorithm implements a differential privacy multi-task learning method, achieving personalized learning through fine-tuning.
    \item \textbf{CENTAUR}: This algorithm realizes personalized federated learning with differential privacy based on FedRep.
\end{itemize}

We select three datasets for this experiment: EMNIST, CIFAR-10, and CIFAR-100. The model used in the experiments is ResNet20 \cite{he2016deep}. In all experiments, the learning rate for the local feature extractor and classifier head is set to 0.01, the learning rate for the server is set to 1, the clipping threshold is 0.01, the model sparsity is set to 0.05, the number of sparsified model layers is 4, the number of training epochs for the local feature extractor is 5, and the number of fine-tuning epochs for the local classifier is 15. For EMNIST, we divide it into 20 clients, and the global model is trained for 50 rounds. For CIFAR-10, we divide it into 20 clients, and the global model is trained for 100 rounds. For CIFAR-100, we divide it into 100 clients, and the global model is trained for 100 rounds.

\subsection{Experimental Results}
\subsubsection{Performance}

\hyperref[table_1]{Table.~\ref*{table_1}} presents the performance of the proposed scheme compared to all baseline schemes. The experiments are conducted on the EMNIST, CIFAR-10, and CIFAR-100 datasets. All datasets follow the basic experimental setup in Section \ref{setting}, with the privacy budget set to 1. The data partitioning method used for each client follows the approach provided by \cite{collins2021exploiting}, ensuring that each client contains data from a specified number (s) of classes. For the EMNIST dataset, the number of classes per client is 5 or 10. For the CIFAR-10 dataset, the number of classes per client is 2, 5, or 10. For the CIFAR-100 dataset, the number of classes per client is 3, 5, or 10.

Based on the data presented in \hyperref[table_1]{Table.~\ref*{table_1}}, our proposed scheme outperforms all state-of-the-art methods in terms of accuracy, achieving improvements of up to 2.9\%. We observe that different baseline methods exhibit varying performance across different data distributions. For instance, on the CIFAR-10 dataset, when s=2, CENTAUR outperforms PPSGD; however, when s=10, CENTAUR performs worse than PPSGD. In contrast, our proposed scheme consistently achieves the best model performance across all data distribution scenarios. This consistency indicates that our method not only surpasses the latest experimental approaches but also adapts more effectively to different data distribution environments.

\begin{table*}[ht]
    \renewcommand{\arraystretch}{1.5}
    \centering
    \caption{Model performance under different sparsity rates.}
    \setlength{\tabcolsep}{7mm}{
    \begin{tabular}{c|c|c|c|c|c|c|c}
        \toprule[1.5pt]
        \rowcolor{lightgray}
        Sparsity Rate & 0.01	&0.05	&0.1	&0.3	&0.5	&0.7	&0.9\\

        \midrule[1pt]
        \multirow{1}{*}{Accuracy} 
        &42.1	&\textbf{42.23}	&41.25	&41.95	&41.32	&42.04	&42.18 \\
    
        \bottomrule[1.5pt]
    \end{tabular}}
    \\ [2mm]
    \label{table_4}
    
\end{table*}

\begin{table*}[ht]
    \renewcommand{\arraystretch}{1.5}
    \centering
    \caption{Model performance under different numbers of reparameterization layers.}
    \setlength{\tabcolsep}{5.9mm}{
    \begin{tabular}{c|c|c|c|c|c|c|c|c}
        \toprule[1.5pt]
        \rowcolor{lightgray}
        Layer Number & 1 & 2 & 3 & 4 & 5 & 6 &7 &8\\

        \midrule[1pt]
        \multirow{1}{*}{Accuracy} 
        &42.23	&41.56	&41.6	&\textbf{43.4}	&42.61	&43.09	&40.74	&41.64  \\
    
        \bottomrule[1.5pt]
    \end{tabular}}
    \\ [2mm]
    \label{table_5}
    
\end{table*}

\begin{table*}[ht]
    \renewcommand{\arraystretch}{1.5}
    \centering
    \caption{Model performance under different hyperparameter $\lambda$.}
    \setlength{\tabcolsep}{7.2mm}{
    \begin{tabular}{c|c|c|c|c|c|c|c}
        \toprule[1.5pt]
        \rowcolor{lightgray}
        $\lambda$ & 0.1	&0.2	&0.3	&0.4	&0.5	&0.6	&0.7\\

        \midrule[1pt]
        \multirow{1}{*}{Accuracy} 
        &43.38	&\textbf{43.92}	&43.41	&42.69	&42.72	&43.06	&43.35 \\
    
        \bottomrule[1.5pt]
    \end{tabular}}
    \\ [2mm]
    \label{table_6}
    
\end{table*}

\subsubsection{Privacy Budget}

We conducted comparative experiments under different privacy budgets using the CIFAR-10, following the experimental setup described in Section \ref{setting}. As shown in \hyperref[table_2]{Table.~\ref*{table_2}}, our proposed scheme consistently outperforms state-of-the-art methods across all privacy budgets. Notably, our method maintains high model accuracy even under low privacy budgets, whereas other baseline methods experience significant drops in accuracy under the same conditions. This indicates that our approach more effectively mitigates the adverse effects of differential privacy, providing superior privacy protection in various federated learning scenarios.

\begin{figure}[hbt!]

\centering
\includegraphics[scale=0.30]{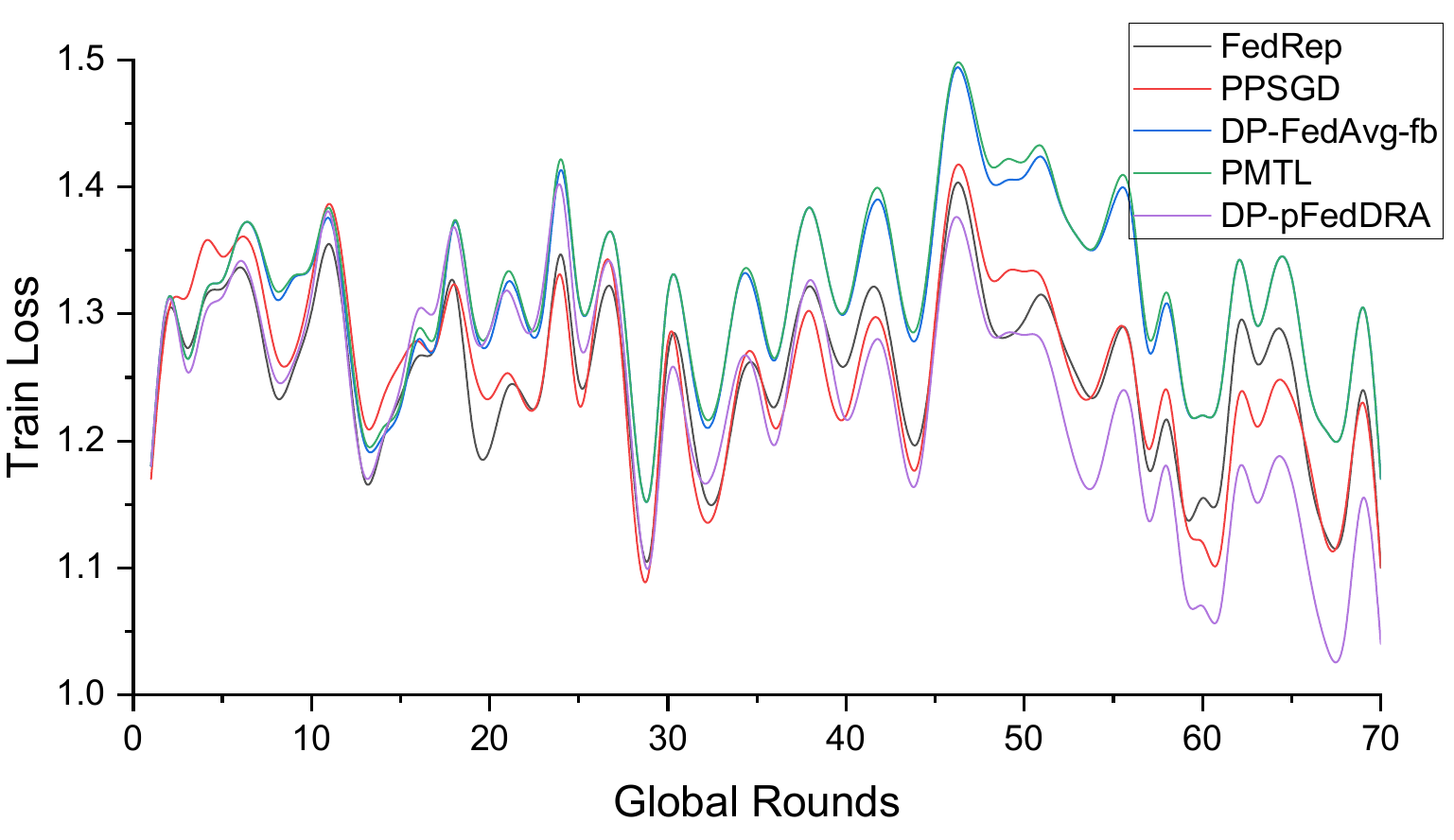}
\caption{Global training loss under different training epochs.}
\label{figure2}
\end{figure}

Furthermore, we record the training loss for different model training methods. Using the CIFAR-10 dataset with a privacy budget of 1, as depicted in \hyperref[figure2]{Fig.~\ref*{figure2}}, the training loss consistently decreases as the number of training rounds increases. Our method achieves lower training loss compared to other approaches, maintaining the lowest loss in the later stages of training. These findings demonstrate that our method better balances the trade-off between differential privacy protection and high model performance.

\subsubsection{Ablation Study}

To demonstrate the impact of the components proposed in this paper on model performance, we conduct experiments on multiple datasets and under different data partitioning scenarios. Here, we select CIFAR-10 and EMNIST, with the privacy budget set to 1. As shown in \hyperref[table_3]{Table.~\ref*{table_3}}, we divide the proposed scheme into two parts: Reparameterization Training (RT) and Dynamic Adaptive Norm (DAN). The numbers in parentheses following the dataset names represent the number of data classes each client has. According to the experimental results, the model performance significantly improves compared to the baseline under Reparameterization Training. When the Dynamic Adaptive Norm is integrated, the model performance is further enhanced. Therefore, both components proposed in this paper effectively improve the integration of differential privacy and personalized federated learning.

\subsubsection{Hyperparameter Study}

The proposed scheme involves the setting of multiple parameters, including the number of model layers undergoing reparameterization, the sparsity rate 
S applied during the sparsification training of the reparameterized model parameters, and the hyperparameter $\lambda$ in the dynamic adaptive norm. Accordingly, this section conducts a series of hyperparameter tuning experiments. As shown in \hyperref[table_4]{Table.~\ref*{table_4}}, the model performance reaches its optimal level when the sparsity rate is set to 0.05. In \hyperref[table_5]{Table.~\ref*{table_5}}, with the sparsity rate set to 0.05, it is evident that the model performance is optimal when the number of reparameterized training layers is 4. Based on the results of these two experiments, we further fine-tune the hyperparameter $\lambda$ in the Dynamic Adaptive Norm. As shown in \hyperref[table_6]{Table.~\ref*{table_6}}, the model performance reaches its optimal level when $\lambda$=0.2. It is evident that through minor adjustments to these parameters, the proposed approach in this study can be effectively tailored to adapt to various scenarios.

\subsection{Conclusion}

In this paper, we address the integration of differential privacy with personalized federated learning by proposing a more efficient differentially private federated learning scheme. Our approach is primarily composed of two components: reparameterization training and dynamic adaptive norm. Reparameterization training identifies important model parameters through local iterative training, reducing the number of parameters and thereby decreasing the amount of noise introduced. The dynamic adaptive norm adjusts the norm of the update information dynamically based on the reparameterization training process, reducing the norm space of the update information and mitigating the impact of clipping  operations on the model. We validate the effectiveness of this method across multiple datasets. Future work may focus on better applying personalized federated learning and differential privacy to larger models.

\section*{DATA AVAILABILITY}
\sloppy
The datasets underlying this article are available at \url{https://biometrics.nist.gov/cs_links/EMNIST/gzip.zip}, \url{https://www.cs.toronto.edu/~kriz/cifar-10-python.tar.gz} and   \url{https://www.cs.toronto.edu/~kriz/cifar-100-python.tar.gz}.

\bibliographystyle{IEEEtran}
\bibliography{mybib}

\vfill

\end{document}